\documentclass{article}

    \PassOptionsToPackage{numbers, compress}{natbib}

\usepackage[preprint]{neurips_data_2023}

\usepackage[utf8]{inputenc} 
\usepackage[T1]{fontenc}    
\usepackage{hyperref}       
\usepackage{url}            
\usepackage{booktabs}       
\usepackage{amsfonts}       
\usepackage{nicefrac}       
\usepackage{microtype}      
\usepackage{xcolor}         
\usepackage{amsmath}
\usepackage{tabularx}
\usepackage{multirow}
\usepackage{graphicx}
\usepackage{algorithm}
\usepackage{algpseudocode}
\usepackage{subcaption}
\usepackage{xspace}
\usepackage{wrapfig}
\usepackage{listings}

\newcolumntype{Y}{>{\centering\arraybackslash}X}
\newcommand{\paratitle}[1]{\vspace{1.5ex}\noindent\textbf{#1}}
\newcommand{\ie}{\emph{i.e.,}\xspace}

\newcommand{\eg}{\emph{e.g.,}\xspace}

\newcommand{\ignore}[1]{}

\title{Evaluating and Improving Tool-Augmented Computation-Intensive Math Reasoning}

\author{
  Beichen Zhang$^{13}$\thanks{Equal contributions.}~,
  Kun Zhou$^{23*}$,
  Xilin Wei$^{4}$,
  Wayne Xin Zhao$^{13}$\thanks{Corresponding author.}~,\\ \textbf{
  Jing Sha$^{5}$,
  Shijin Wang$^{56}$,
  Ji-Rong Wen$^{123}$}
  \\
  $^1$Gaoling School of Artificial Intelligence, Renmin University of China.\\
  $^2$School of Information, Renmin University of China.\\
  $^3$Beijing Key Laboratory of Big Data Management and Analysis Methods.\\
  $^4$College of Computer Science, Sichuan University.\\
  $^5$iFLYTEK Research.\\
  $^6$iFLYTEK AI Research (Central China).\\
  \texttt{\{zhangbeichen724,wiselnn570,batmanfly\}@gmail.com},~\texttt{francis\_kun\_zhou@163.com}\\
  \texttt{\{jingsha,sjwang3\}@iflytek.com},\texttt{jrwen@ruc.edu.cn}
}

\begin{document}

\maketitle

\begin{abstract}
Chain-of-thought prompting~(CoT) and tool augmentation have been validated in recent work as effective practices for improving large language models~(LLMs) to perform step-by-step reasoning on complex math-related tasks.
However, most existing math reasoning datasets may be not able to fully evaluate and analyze the ability of LLMs in manipulating tools and performing reasoning, as they may only require very few invocations of tools or miss annotations for evaluating intermediate reasoning steps.
To address the issue, we construct \textbf{CARP}, a new Chinese dataset consisting of 4,886 computation-intensive algebra problems with formulated annotations on intermediate steps.
In CARP, we test four LLMs with CoT prompting, and find that they are all prone to make mistakes at the early steps of the solution, leading to wrong answers.
Based on this finding, we propose a new approach that can deliberate the reasoning steps with tool interfaces, namely \textbf{DELI}.
In DELI, we first initialize a step-by-step solution based on retrieved exemplars, then iterate two deliberation procedures that check and refine the intermediate steps of the generated solution, from the perspectives of tool manipulation and natural language reasoning, until obtaining converged solutions or reaching the maximum turn.
Experimental results on CARP and six other datasets show that the proposed DELI mostly outperforms competitive baselines, and can further boost the performance of existing CoT methods.
Our data and code are available in \url{https://github.com/RUCAIBox/CARP}.
\end{abstract}

\section{Introduction}
Recently, large language models~(LLMs) (\eg GPT-3 and ChatGPT) have shown remarkable zero-shot and few-shot performance on various tasks~\cite{brown2020language,Ouyang2022TrainingLM,Zhao2023ASO}, including language generation and question answering.
As LLMs have been pre-trained on a large amount of text data, covering broad types of world knowledge, existing work also shows that LLMs can solve complex tasks, \eg math reasoning~\cite{Wei2022ChainOT} and college entrance exam~\cite{OpenAI2023GPT4TR,Zhong2023AGIEvalAH}.

To evaluate the capacity of LLMs for solving complex tasks, math reasoning datasets have been widely used as testbeds, \eg GSM8K~\cite{cobbe2021training} and MATH~\cite{Hendrycks2021MeasuringMP}, where the math problems can not be directly answered but require multi-step reasoning.
To elicit LLMs for step-by-step reasoning, chain-of-thought~(CoT)~\cite{Wei2022ChainOT,Chen2023ChatCoTTC} has become the de facto prompting strategy, where LLMs can be guided to generate a solution consisting of a series of intermediate steps for reaching the answer.
However, previous work also reveals that LLMs are prone to make mistake at intermediate steps, especially for numerical computation~\cite{Qian-arxiv-2022-Limitations,Lu-arxiv-2022-Survey,Frieder-arxiv-2023-Math,Yuan2023HowWD,Zhao2022JiuZhangAC}, and yet a minor mistake would lead to a totally wrong answer.
To alleviate it, a line of work~\cite{parisi2022talm,gao2022pal,Chen2022ProgramOT,schick2023toolformer,LYU2023FaithfulCR,Joy2023Symbol,Press2022MeasuringAN,Yao2022ReActSR,Jiang2023StructGPTAG,Gou2023CRITICLL,Imani2023MathPrompterMR} employs external tools to make up for the weakness of LLMs, and can greatly improve the answer accuracy on math reasoning tasks.
%
With the rapidly evolving LLMs and tool-augmented methods, it is necessary to adopt a suitable math reasoning dataset for evaluating them systematically and differentially.
Whereas, the problems in most existing math reasoning datasets may only require the one-off utilization of tools~\cite{gao2022pal,Chen2022ProgramOT,LYU2023FaithfulCR,Joy2023Symbol}, which are not adequate to fully measure the ability of tool manipulation in existing methods.
Besides, although the wrong answers mostly derive from the incorrect intermediate steps in step-by-step reasoning, most existing datasets can not be utilized for testing that, due to lack of formal annotations of the intermediate steps in the solution text.
The two issues limit existing math reasoning datasets to systemically evaluate and analyze LLMs and tool-augmented methods.

To address them, we construct a new Chinese dataset that consists of 4,886 \textbf{C}omputation-intensive \textbf{A}lgeb\textbf{R}a \textbf{P}roblems associated with formulated annotations of all the intermediate steps, namely \textbf{CARP}.
In CARP, all the problems require deriving multiple intermediate math expressions based on math knowledge, and solving them based on arithmetical knowledge, which make it a complex and difficult dataset to evaluate the computation-intensive math reasoning ability.
In addition, the formulated annotations also enable researchers to test the accuracy of intermediate reasoning steps for analyzing the errors of LLMs.
As shown in Table~\ref{tab:ExpAcc}, four popular LLMs with CoT prompting can not solve over half of the problems in our CARP, indicating the difficulty of CARP.
Furthermore, we also find that all LLMs are more likely to make mistakes in the first step (over 69\%), leading to totally wrong solutions and answers.
It reveals that LLMs mostly fail in performing early reasoning steps, and can not correct the errors in the latter steps.
Based on CARP, we also devise a variety of fine-grained interfaces based on available tools, to provide practical functionalities for handling complicated calculations.
These interfaces can also be applied to other math reasoning datasets to improve the tool manipulation capacity of LLMs.

Considering that LLMs can not fix the errors in early steps by themselves, we propose a new approach that can \textbf{deli}berate the reasoning steps of LLMs with interfaces of tools, namely \textbf{DELI}.
In DELI, we first initialize a step-by-step solution for the given question based on retrieved relevant exemplars, then iterate two deliberation procedures that check and refine the generated step-by-step solution from the perspectives of tool manipulation and natural language reasoning, until reaching the stop condition, \eg solution has converged or iterations reach the maximum number.
Such a way is similar to the solution checking process of humans, and can elicit LLMs to deliberate and correct the possible errors in intermediate steps of the solution.
We evaluate our proposed DELI and existing prompting methods on CARP and six other computation-intensive datasets.
Experimental results show that the proposed DELI mostly outperforms competitive baselines (\eg 9.35\% accuracy improvement over the best baseline on CARP), and can further boost the performance of existing CoT prompting methods.

To summarize, our major contributions are: 

$\bullet$ We construct a new dataset named CARP with formulated annotation of intermediate reasoning steps for systematically evaluating LLMs in solving computation-intensive math problems, and devise interfaces with practical functionalities to help LLMs.

$\bullet$ We propose DELI, a new approach that can deliberate and correct the reasoning steps of LLMs with interfaces of tools.

$\bullet$ We conduct extensive experiments to show the superiority of our DELI over existing prompting methods on 7 computation-intensive math reasoning datasets.

\begin{table}[t]
\begin{minipage}{0.5\linewidth}
\setlength{\tabcolsep}{4pt}
    \centering
    \caption{Statistics for CARP dataset.}
    \label{tab:CARP}
    \begin{tabular}{lr}
    \toprule
    \textbf{Statistic}                         & \textbf{Number}  \\
    \midrule
    \# of training samples & 3,410 \\
    \# of development samples & 500 \\
    \# of testing samples &  976 \\
    \midrule
    \# of nodes (Avg./Max) & 6.0/18 \\
    \# of edges (Avg./Max) & 5.7/25\\
    \# of expression nodes (Avg./Max) & 4.7/15 \\
    \midrule
    Problem length (Avg./Max) & 52.1/257 \\
    Solution length (Avg./Max) & 71.3/278 \\
    \bottomrule             
    \end{tabular}
\end{minipage}
\begin{minipage}{0.5\linewidth}
    \centering
    \includegraphics[width=0.95\linewidth]{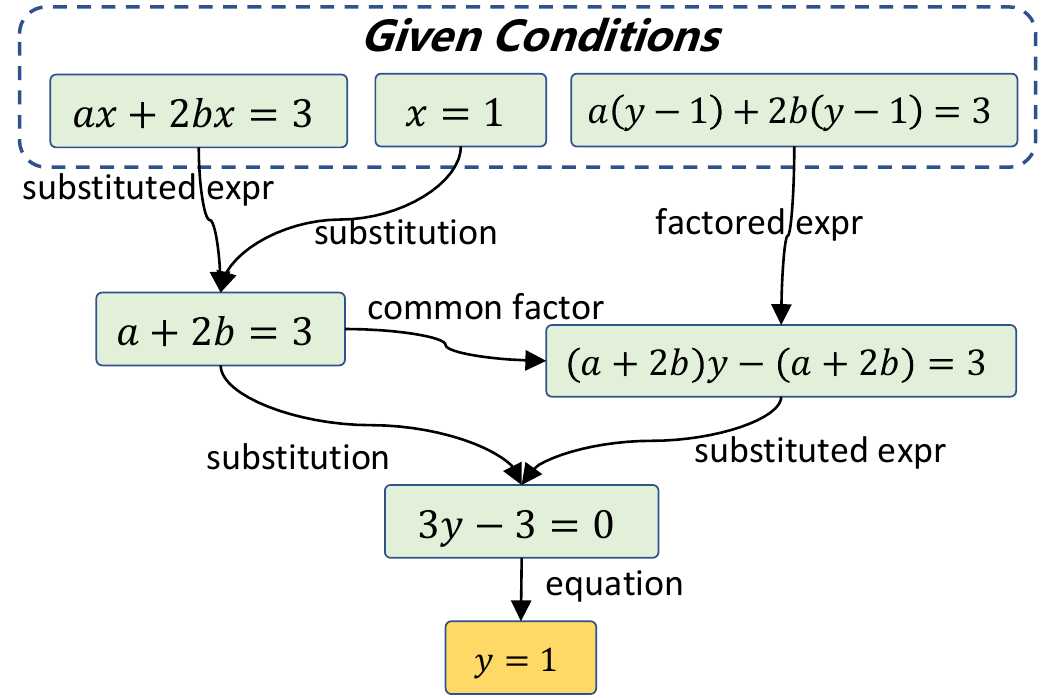}
    \captionof{figure}{An EFG annotation example for CARP.}
    \vspace{-1em}
    \label{fig:graph}
\end{minipage}

\end{table}

\section{CARP Dataset}
Computation-intensive math reasoning task aims to solve complex math problems that require performing multi-step arithmetical computation and reasoning based on mathematical knowledge.
Typically, to solve a computation-intensive math problem, humans or models need to iterate the process multiple times that derives the formula and then compute the result (via calculators or other tools), until obtaining the final answer.
In this way, the accuracy of intermediate reasoning and computation steps is crucial, where a subtle error would lead to totally wrong answers.
In this paper, we construct a new dataset CARP~(Computation-intensive AlgebRa Problems) that provides the formulated annotations of all the intermediate steps for the computation-intensive middle school math problems.
Based on the annotations, we also design a set of interfaces with fine-grained computation functions, to help LLMs manipulate commonly-used tools for solving these problems.

\subsection{Dataset Construction}
Although there are a number of computation-intensive math problems in available datasets, their solutions are generally in not well-formulated natural language and may omit intermediate steps~\cite{hosseini2014learning,koncel2015parsing,roy-roth-2015-solving,ling2017program,Patel2021AreNM,cobbe2021training,Hendrycks2021MeasuringMP}.
To construct a well-formulated dataset, we first collect real-world computation-intensive math problems, then invite crowd-sourced workers to extract and annotate their expression flow graph.

\paratitle{Data Collection.}
We collect the math problems and their step-by-step solutions from a Chinese education website Zhixue\footnote{https://www.zhixue.com/}, which contains vast problems to provide education assistance for students.
We mainly crawl middle school math problems, since they are of moderate difficulty and require basic arithmetical computations (\eg quadratic equation) and mathematical knowledge (\eg Veda's theorem), making them a good testbed for computation-intensive math reasoning.
We first crawl about 1,000,000 problems with solutions.
Then, to obtain computation-intensive problems, we design hand-crafted rules based on SymPy to roughly extract and count the computation steps in solutions, and only select the ones with both over one computation step and over two reasoning steps.
Finally, we invite math teachers to select about 10,000 high-quality examples for annotation.

\paratitle{Expression Flow Graph Annotation.}
In a math problem, the natural language solution can be generally formulated as a directed acyclic graph~(DAG), where the nodes and edges refer to the intermediate results and derivation steps, respectively~\cite{Jahring2020TheEO}.
For computation-intensive problems, we consider a special DAG format that adopts intermediate math expressions as nodes.
We name it \emph{expression flow graphs}~(EFG), as it can explicitly show how to derive new math expressions based on existing ones in the step-by-step reasoning process.
In this way, a solution can be formulated as: starting from initial condition nodes within the problem, we continue deriving new nodes (\ie intermediate math expressions) from existing nodes, until reaching the final expression that can obtain the answer, where the computation results of intermediate expressions can be utilized for evaluation.
Whereas, a math problem may involve special initial conditions that are hard to be converted into readable expressions, \eg Equations have rational solutions. 
Thus, we add a special type of node to store these conditions in natural language, while guaranteeing that all the derived new nodes are math expressions.
As an example, the EFG annotation of the Problem in Table~\ref{tab:example} is shown in Figure~\ref{fig:graph}.

Based on the above definition, we invite five middle school math teachers as crowd-sourced workers to annotate the formulated EFGs of our collected problems.
The annotation process is similar to the information extraction process~\cite{Chang2006ASO}, where we first extract the nodes and then link them to compose the graph.
Concretely, we first rely on hand-crafted rules to automatically extract the math expressions and text conditions from the solution texts as node candidates.
Then, we ask math teachers to link the related node candidates and annotate their corresponding relations.
To reduce the difficulty, we utilize heuristic rules to select the most possible related nodes and relations as references.
Consequently, we can collect a set of edges with special relations connecting several nodes from the node candidates, which compose the EFG of a problem.
After annotation, we further design an automatic verification program to verify the completeness of the EFG and the validity of relations, and filter improper ones.
Besides, we also ask teachers to check the annotated EFGs from each other, to judge if the EFG has fully covered the whole problem-solving process of the problem, and refine the incomplete ones.

\subsection{Dataset Details}

\paratitle{Dataset Description.}
The statistics of the CARP dataset are shown in Table~\ref{tab:CARP}. 
CARP consists of 4,886 middle school computation-intensive algebra problems, and each problem is associated with a natural language solution and an annotated EFG.
Our annotated EFG explicitly depicts the step-by-step reasoning process of a math problem in a readable and concise format.
On average, an EFG contains 6.0 nodes and 5.7 edges, as we only keep the expressions and conditions that lead to the final answer in the EFG.
Besides, an EFG has 4.7 expression nodes on average, which are the main stem of the whole reasoning process and can be used for evaluating the accuracy of intermediate steps.

To solve the problems in CARP, LLMs require to iteratively perform reasoning based on math knowledge to correctly derive the intermediate math expressions, and solve it accurately.
As the example in Table~\ref{tab:example}, given the conditions, a reasonable solving process should first deduce the intermediate equation $a+2b=3$ by substituting $x=1$ into $ax+2bx=3$, and then reformulate the equation $a(y-1)+2b(y-1)=3$ to support plugging $a+2b=3$ into it.
Such a reformulation step is not easy to reason out, and ChatGPT has made a mistake there, leading to a wrong answer.

\begin{table}[]
    \centering
    \caption{An example from the CARP dataset, which is translated into English. Errors are annotated with red color.}
    \begin{tabular}{l p{11cm}}
    \toprule
      \textbf{Problem} & The solution to the equation $ax + 2 bx = 3$ is $x = 1$ , then the solution to the equation $a ( y - 1 ) + 2 b ( y - 1 ) = 3$ is ? \\
      \midrule
      \textbf{Solution} & From the question we have : $a + 2 b = 3$ , $a ( y - 1 ) + 2 b ( y - 1 ) = 3$ . Rectifying gives $( a + 2 b ) y - ( a + 2 b ) = 3$ , \emph{i.e.,} $3 y - 3 = 3$ , therefore $y = 2$ . \\
      \addlinespace[0.2em]
    \textbf{ChatGPT} & Substituting $x = 1$ into $ax + 2bx = 3$ gives $a + 2b = 3$ , and \textcolor{red}{substituting $y - 1$ gives $a ( y - 1 ) + 2 b ( y - 1 ) = 3$}, which simplifies to \textcolor{red}{$ay + by = 3$} . $\cdots\cdots$ The answer is $y = \frac { 3 } { 3 - a } - 1$\\ 
    \addlinespace[0.2em]
    \textbf{Error Type} & Reasoning error. \\
    \bottomrule
    \end{tabular}
    
    \label{tab:example}
\end{table}

\label{sec:metric}

\paratitle{Evaluation Metrics.}
Based on EFGs, we can evaluate the intermediate step-by-step reasoning process of LLMs.
Specifically, we propose two new metrics, \ie \emph{ExpAcc} and \emph{Fail@where}.
ExpAcc measures the accuracy rate of the intermediate expressions of a problem in the generated output.
Considering that a math problem may have different ways to solve it, we also regard the ancestors of a correct intermediate expression as true ones, as they derive the right expression.
In this way, ExpAcc can be obtained by first finding accurately matched expression nodes and then counting their ancestors and themselves as accurate ones for computing the rate.
We leverage SymPy to judge if two math expressions are matched.
Fail@where is another type of metric for analyzing where are the causes of incorrect answers, and we define three implementations, \ie Fail@first, Fail@middle, and Fail@last.
The three metrics refer to the rates of making the first mistakes in the first step, middle steps, and last step (before the answer) within all generated incorrect solutions, respectively.

\begin{table}[]
    \centering
    \caption{Evaluation results of different LLMs with CoT prompting on CARP.}
    \label{tab:ExpAcc}
    \begin{tabular}{lccccc}
        \toprule
        \multirow{2}{*}{\textbf{Models}} &  \multirow{2}{*}{\textbf{Acc.}} &  \multirow{2}{*}{\textbf{ExpAcc}} & \multicolumn{3}{c}{\textbf{Fail@where}} \\
        \cline{4-6}
        \addlinespace[0.2em]
        & & & \textbf{Fail@first} & \textbf{Fail@middle}  & \textbf{Fail@last}   \\
        \midrule
        \texttt{text-davinci-002} & 31.15 & 37.45 & 79.04 & 11.29 & 9.65 \\
        \texttt{text-davinci-003} & 37.50 & 44.89 & 73.61 & 15.41 & 10.98 \\
        \texttt{claude-v1.3} & 40.78 & 46.89 & 76.85 & 12.08 & 11.05 \\
        \texttt{gpt-3.5-turbo} & \textbf{49.39} & \textbf{56.48} & 69.69 & 16.36 & 13.94 \\
        \bottomrule
    \end{tabular}
\end{table}

As shown in Table~\ref{tab:ExpAcc}, we evaluate competitive LLMs on CARP with chain-of-thought prompt~\cite{Wei2022ChainOT} and report the answer accuracy, ExpAcc, and Fail@where.
First, all LLMs can not solve over half of the problems in CARP (under 50.0), and the accuracy of intermediate steps is relatively lower (under 57.0), indicating the difficulty of computation-intensive math reasoning.
Second, all LLMs are more likely to make mistakes in the first step, while less likely in the last step. 
It demonstrates that LLMs are prone to fail in early steps, due to misuse of improper math knowledge or wrong calculations.
Thus, careful deliberations on early steps might be promising to reduce errors of the model.

\subsection{Tool Interfaces}
\label{sec:tool}
As the results in Table~\ref{tab:ExpAcc} and existing work~\cite{Hendrycks2021MeasuringMP,Lu-arxiv-2022-Survey,Frieder-arxiv-2023-Math,Yuan2023HowWD}., it is hard for LLMs to solve computation-intensive math problems, especially for numerical calculation.
In the real world, humans can utilize tools (\eg calculator) to avoid errors in manual work.
Inspired by it, we consider augmenting LLMs with tools for handling complicated calculations.
Considering the complexity of math calculation, we devise multiple interfaces based on available tools, to provide specific and practical functionalities.
All the interfaces are formulated into a unified format with detailed descriptions, to support convenient manipulation of LLMs.
Concretely, we mainly utilize SymPy~\cite{Meurer2017SymPySC} as the tool, which is a Python library including various basic and advanced arithmetic operators.
Based on it, we encapsulate three types of interfaces to help the computation of LLMs:
(1) \textbf{Numerical Computation}: compute the value $v$ of an expression $e$ by calculating directly or substituting existing conditions. 
(2) \textbf{Equation Solving}: solve an equation or inequation $e$, or solve the system of equations or inequalities $\{e\}$.
(3) \textbf{Expression Transformation}: transform an expression $e$ into the desired format $e'$.

Based on them, we devise fine-grained interfaces covering commonly-used functionalities in math calculation.
We set the name, arguments, and output formats of each interface, associated with a docstring that provides a natural language explanation for its usage.
These interfaces are general to various computation-intensive math reasoning tasks, and can help LLMs perform complex computations.
In addition, we also add a special interface,  \emph{think}, which can utilize the LLM to analyze existing conditions, deduce new conclusions, and create new math expressions, before or after tool manipulation.
It can also help handle the cases that fail to invoke computation interfaces, where LLMs \emph{think} to produce an output instead, to prevent the solving process from being interrupted.

\subsection{Dataset Discussion}
Our proposed CARP dataset focuses on systematically evaluating LLMs in solving computation-intensive math problems.
CARP exhibits three key characteristics:
First, solving problems in CARP involves multi-step reasoning with math domain knowledge, math expression understanding, and complex computations;
Second, fine-grained interfaces are provided in CARP for LLMs to evaluate the tool manipulation ability in complex reasoning.
In this scenario, LLMs should understand the usage of fine-grained interfaces, and invoke them reasonably multiple times based on reasoning and mathematical knowledge in the solving process.
Third, evaluation metrics for intermediate reasoning steps based on formulated annotations are employed to better analyze the multi-step reasoning performance of LLMs, while existing datasets mainly focus on evaluating the outcome accuracy of solutions~\cite{Zhong2023AGIEvalAH,cobbe2021training,Hendrycks2021MeasuringMP}.
Via those metrics, researchers can quantify the models' mastery of the solution process and thus acquire more clues for improving models.
\section{Approach}
\begin{figure*}[t]
    \centering
    \includegraphics[width=\textwidth]{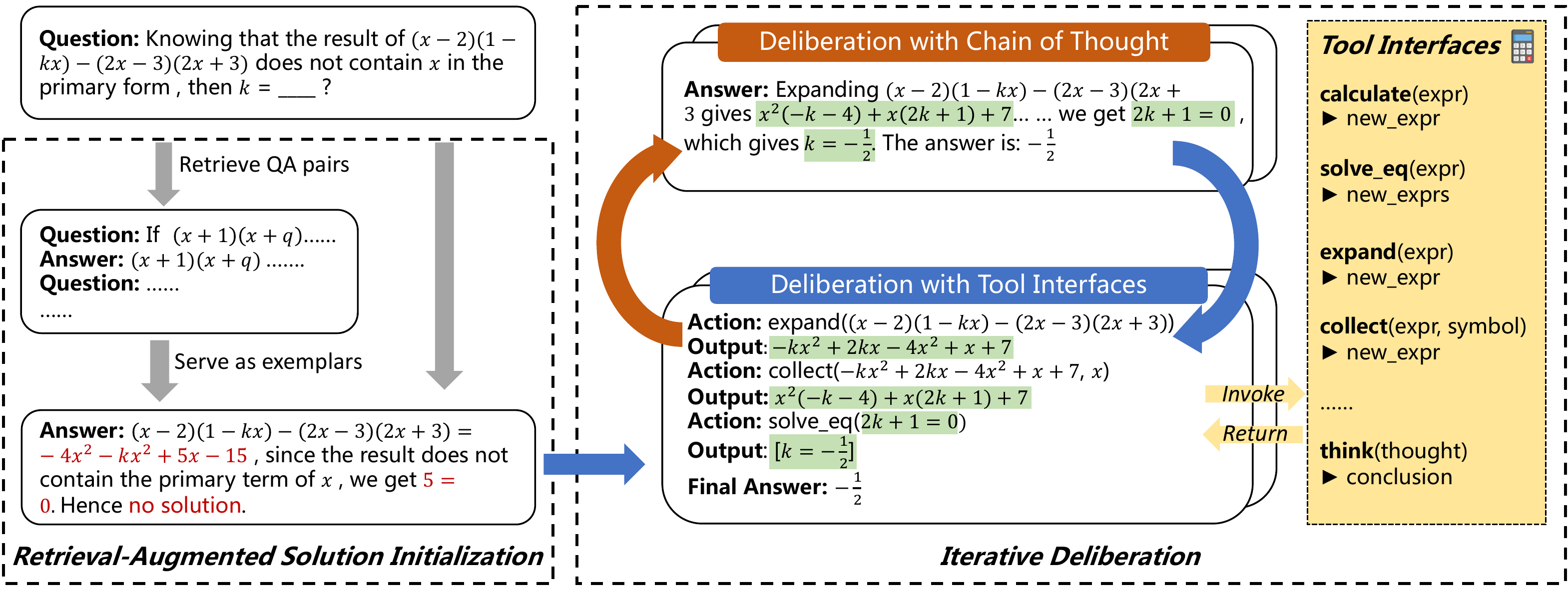}
    \caption{The overview of our DELI. DELI first initializes the step-by-step solution via retrieval-augmented strategy, and then performs iterative deliberation with tool manipulation and chain of thought, respectively.}
    \label{fig:model}
\end{figure*}

According to the results in Section~\ref{sec:metric}, it is hard for LLMs to solely solve computation-intensive math problems, and they often make mistakes in early reasoning steps.
Inspired by humans that often review and check the solutions, we propose a new approach that can \textbf{deli}berate the reasoning steps of LLMs with interfaces of tools, namely \textbf{DELI}.
The overview of DELI is shown in Figure~\ref{fig:model}.
In DELI, we leverage a retrieval-augmented chain-of-though prompting strategy to initialize a step-by-step natural language solution.
Then, we iterate the two-stage deliberation method that checks and refines the solution from the perspectives of natural language reasoning and tool manipulation.
After multiple iterations, we can finally obtain a more reasonable solution with the answer.

\subsection{Retrieval-Augmented Solution Initialization}
As our approach focuses on performing deliberation on the generated solution of LLMs, we aim to first initialize a high-quality step-by-step solution for the given question that covers useful mathematical knowledge and arithmetic operators.
Therefore, we propose to retrieve relevant problems and solutions as the exemplars, and then utilize the chain-of-thought prompting method~\cite{Wei2022ChainOT} to generate the initial solution based on them.
Concretely, given a math problem $p$, we first utilize a retriever to select the top-$k$ relevant problems $C=\{ \langle p_i, s_i \rangle \}^{k}_{i=1}$ from the candidate pool based on question-question matching, where the retriever can be either lexicon-based~\cite{Robertson2009ThePR} or dense retrieval models~\cite{Gao2021SimCSESC}.
Then, the retrieved problems with their associated step-by-step solutions, will be employed to compose the input prompt, to elicit LLMs for performing chain-of-thought reasoning.
The pattern of the input prompt is denoted as:
``\emph{You are a helpful assistant for solving math problems in LaTeX format: [$p_1$], [$s_1$], $\cdots$, [$p_k$], [$s_k$], [$p$]}''.
In this way, LLMs would follow the in-context exemplars to perform step-by-step reasoning, and can also refer to useful mathematical knowledge and arithmetic operators from them, leading to more high-quality initial solutions for deliberation.
Note that such a way also supports other prompting methods to initialize solutions.

\subsection{Iterative Deliberation}
Based on the initial solution of the given question, we iterate two types of deliberation procedures, \ie deliberation with tool interfaces and deliberation with chain of thought, until reaching the stop condition.
In the two deliberation procedures, we adopt specific in-context exemplars to guide LLMs, for checking and correcting the errors in the current solution.
Next, we first introduce the details of the two deliberation procedures, and then present the stop condition.
The algorithm of the iterative framework is illustrated in the supplemental materials.

\paratitle{Deliberation with Tool Manipulation.}
Since LLMs are prone to make mistakes in numerical calculation, we design the procedure of deliberation with tool manipulation, for seeking help from external tools to address it.
Based on our devised interfaces in Section~\ref{sec:tool}, we aim to rewrite the current solution into a process that orderly invokes the interfaces to produce the result.
In this way, the deliberation procedure is divided into a sequence of steps, where the LLM should select the interface and then invoke it to produce the intermediate result in each step.

Concretely, first, we construct an instruction that introduces the goal and formats of this procedure, and the details of all available interfaces.
For each interface, we not only list its name, arguments and description, but also provide an example to exhibit the way to use it, \eg ``\emph{expand(expression: str)$\rightarrow$ new expression: str: Expand the expression into a polynomial}. \eg expand($(x + 1) ^ 2$) -> $x ^ 2 + 2x + 1$''.
Then, we demonstrate several exemplars to guide LLMs to invoke the interfaces.
Each exemplar consists of four parts, \ie a question, a trial, multiple actions, and their outputs.
The trial is the initial step-by-step solution to the given question, which may contain a few errors requiring correction.
Actions are a series of interface invocation operations derived from the trial, and outputs are the intermediate results by executing the actions, \eg ``\emph{Action: solve\_eq($2k+1=0$). Output: [$k=-\frac{1}{2}$]}''.
Based on the instruction and exemplars, the LLM would be elicited to generate the action in formal language iteratively (\ie selecting the interface and setting its arguments), then execute it to obtain the intermediate result, until reaching the answer.
To guarantee the continuity of the deliberation procedure, we set a special token after the generated action, for pausing the generation process and waiting for the result of interface invocation.
In the iterative selection-then-execution process, we can deliberate the intermediate steps of the generated solution, and benefit from tool manipulation for accurate numerical computation.

\paratitle{Deliberation with Chain of Thought.}
After deliberation with tools, we can obtain the solution in formal language consisting of a series of actions to invoke interfaces and their outputs.
Next, we further deliberate the solution with chain of thought to reorganize it into the natural language format, which can better make use of the learned textual knowledge from LLMs to recheck it and also improve the readability.

Similarly, we also leverage an instruction with in-context exemplars to compose the input prompt.
The instruction is ``\emph{You have access to both natural language problem solving processes and formal problem solving processes, but there may be errors within them. You need to learn the correct methods in order to better solve problems. }'', to introduce the goal of the deliberation procedure.
All the exemplars are composed of four components, \ie a question, a given solution, the verification, and the revised solution.
The given solution is the last natural language solution that is either the initial solution or the solution from the last deliberation iteration with chain of thought, and the verification is the formal language solution from the last deliberation procedure with tool interfaces.
The revised solution is the result of integrating the two types of solutions into the chain-of-thought manner, where the errors and unreasonable steps have been corrected.
Guided by the exemplars, LLMs would also deliberate the intermediate steps from in-context solutions, and generate a new natural language solution to answer the given problem. 
Besides, as there are often inconsistent intermediate computation results in the in-context solutions, we also add an instruction to elicit LLMs to trust more on the result from tool manipulation, \ie ``\emph{If the computed result in the verification differs from the computed result in the given solution, the computed result in the verification must be used as the standard}''.

\paratitle{Stop Conditions of Iteration.}
The devised two deliberation procedures would be alternated multiple times, where the solution might be iteratively revised and improved.
To control the cost, we set the stop conditions of the iteration process.
First, once the solution of the new iteration is the same as the last one, the iteration will be stopped, since it reflects that the iteration has converged.
Second, if the answers of the two deliberation procedures are consistent, we will also stop the iteration.
Third, if we have reached the maximum number of iterations, the answer from the last deliberation procedure with tool manipulation will be regarded as the final answer, as such a procedure can better solve computation subproblems, leading to a more accurate answer.
\begin{table}[h]
\centering
\caption{Basic information about datasets in evaluated datasets. MS and HS refer to ``middle school'' and ``high school'', respectively.}
\begin{tabularx}{\textwidth}{XYccccc}
\toprule
\textbf{Dataset} &
  \textbf{Source} &
  \textbf{Language} &
  \textbf{Domain} &
  \textbf{Difficulty} &
  \textbf{Train} &
  \textbf{Test} \\
\midrule
CARP           & Ours    & Chinese      & Algebra          & MS          & 3,410          & 976       \\
Algebra          &  MATH   & English       & Algebra      & HS    & 1,744          & 1,187            \\
Prealgebra    & MATH & English  & Algebra          & HS         & 1,205          & 871         \\
Count. \& Prob.    & MATH & English  & Probability          & HS         &  771          & 474          \\
Num. Theory    & MATH & English  & Number Theory          & HS         &  869          & 540          \\
GK-Cloze   & AGIEval & Chinese  & Mixture          & HS         &  -          & 220          \\
SAT-Math    & AGIEval & English  & Mixture          & HS         &  -          & 351          \\
\bottomrule
\end{tabularx}
\label{tab:benchmark}
\end{table}

\section{Experiment}

\subsection{Main Experiments}

\paragraph{Settings.}
In addition to our CARP dataset, we also collect six existing computation-intensive math problem datasets for evaluation, including Algebra, Prealgebra, Counting and Probability~(CP), and Number Theory~(NT) from MATH benchmark~\cite{Hendrycks2021MeasuringMP} and GK-Math-Cloze~(GKC) and SAT-Math from AGIEval~\cite{Zhong2023AGIEvalAH}.
The statistics of these datasets are shown in Table~\ref{tab:benchmark}
These datasets need multi-step reasoning and computation to solve the problem, and the required knowledge varies from middle school to math competitions.
We also show the details of baselines and implementation in supplementary material.

\begin{table}[t]
\centering
\caption{Results on 7 computation-intensive math reasoning datasets. We copy results of LP from~\citet{Guo2023LearningTP}. The best and second-best methods are marked in bold and underlined respectively. }.
\begin{tabularx}{\textwidth}{Xccccccccc}
\toprule
\textbf{Methods} &
  \textbf{CARP} &
  \textbf{Algebra} &
  \textbf{Prealgebra} &
  \textbf{CP} &
  \textbf{NT} &
  \textbf{GKC} &
  \textbf{SAT} &
  \textbf{Avg.} \\ 
\midrule
Random CoT           & 49.39          & 49.37          & 55.57          & 32.91          & 29.81          & 14.41      & 65.91 & 42.48 \\    
Complex CoT          &    48.06        & 51.64          & 53.73          & 32.91          & 32.22          & -             & -   & -  \\
Retrieval CoT        & 63.93          & 53.75          & 56.72          & 33.12          & 30.00             & -         & -   & -  \\
\midrule
PAL                  &      40.00          &     34.29      &     50.52      &      35.86     &    31.30       &       5.93        &   47.73 & 35.09    \\
ReAct           & \underline{64.11}          & \underline{54.51}          & 54.53          & \textbf{41.77}          & 31.67          & \underline{16.94}         & \underline{72.27} & \underline{48.07} \\
\midrule
LP                   & -              & 49.60          & 52.30          & 30.20          & 29.80          & -             & -  & -   \\
PHP                  &       61.68         & 54.42          & \underline{57.86}          & 36.71          & \textbf{35.37}          &     \underline{16.94}          &   71.82   & 47.82 \\
\midrule
Iterative CoT                  &        61.27        &     52.74      &     55.34      &     33.97      &      29.81     &      14.41         &    69.55   &   45.30   \\
Iterative ReAct          &        61.17        &     53.92    &     52.12     &   37.34        &         32.22  &       15.25      &    70.00  & 46.00 \\
\midrule
DELI & \textbf{73.46} & \textbf{59.65} & \textbf{58.32}          & \underline{39.03} & \underline{33.15}          &  \textbf{17.80}             &   \textbf{74.54} & \textbf{50.85}  \\ 
\bottomrule
\end{tabularx}
\label{tab:main}
\end{table}

\paratitle{Main Results.}
Table~\ref{tab:main} compares DELI with other baselines on the 7 datasets.
For the comparison between CoT prompting methods, Retrieval CoT outperforms Random CoT on average, indicating that reviewing relevant knowledge and problem-solving idea benefit answering complex math problems.
Augmented with our provided interfaces, ReAct achieves better average performance than CoT prompting methods, demonstrating the effectiveness of wrapping tools to aid math reasoning through interfaces.
Besides, in the comparison between tool-augmented prompting, the performance of ReAct is better than PAL on the benchmark.
It indicates that reasoning with intermediate results is important for solving computation-intensive math problems.
Finally, our proposed DELI performs better than competitive baselines in most cases.
DELI improves upon the basic components of deliberations, including CoT prompting methods and ReAct, while iterative variants (\ie Iterative CoT and Iterative ReAct) without our designed deliberations underperform basic solutions.

\subsection{Analysis}

\begin{figure*}[t]
    \centering
    \includegraphics[width=\textwidth]{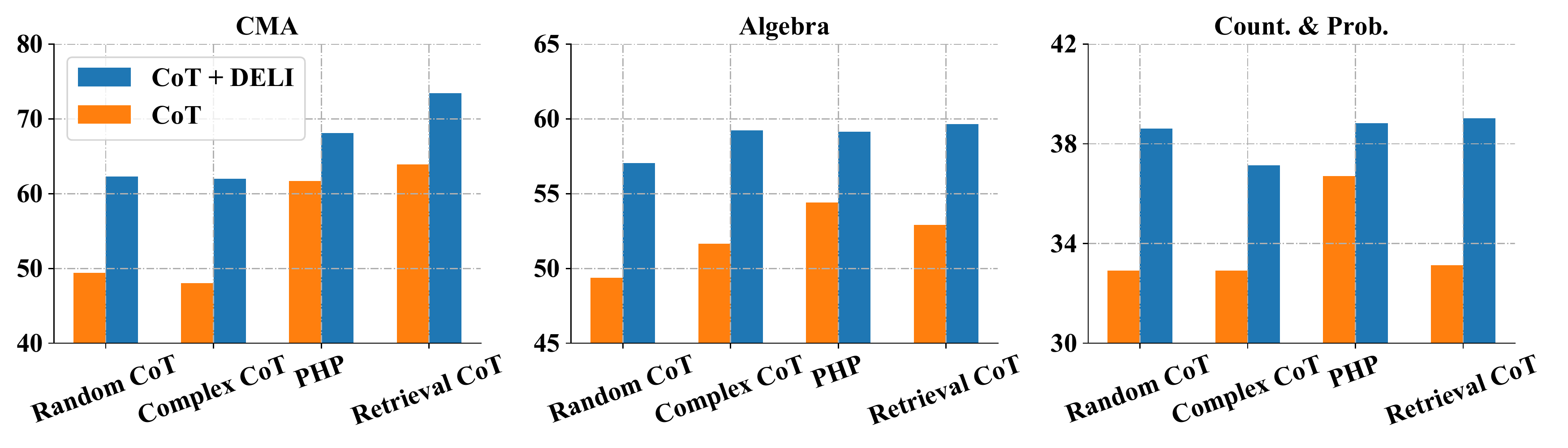}
    \caption{The results of combining DELI with existing CoT methods.}
    \label{fig:improve}
\end{figure*}

\paratitle{Combining with Existing CoT methods.}
\label{sec:improve}
In DELI, we initialize a step-by-step solution by retrieving problems and solutions as exemplars.
However, it is noted that DELI also supports other prompting methods to initialize solutions.
We report the performance of DELI combined with different CoT methods on CMA, Algebra, and Count. \& Prob. datasets.
As shown in Figure~\ref{fig:improve}, DELI brings improvement upon all CoT methods, which demonstrates our framework can be applied to various CoT prompting methods.
In particular, DELI can further boost the performance of the existing iterative method PHP, which shows that incorporating fine-grained interfaces to assist reasoning can find and fix errors that are difficult to correct with CoT, such as complex calculation errors.

\begin{figure*}[t]
    \centering
    \includegraphics[width=\textwidth]{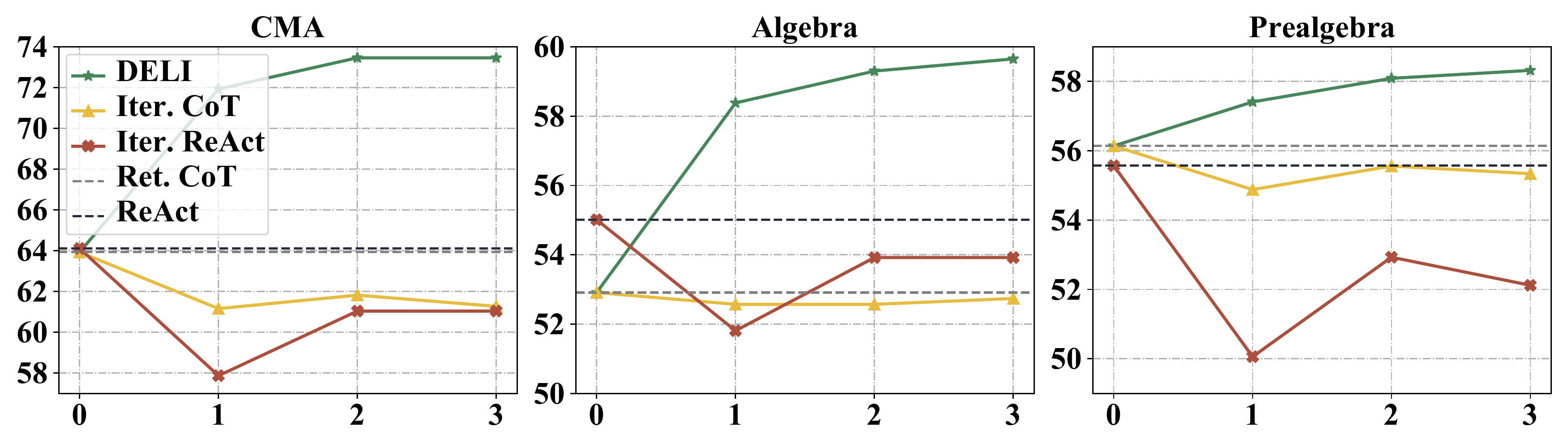}
    \caption{Accuracy of different methods w.r.t. the maximum number of iteration turns.}
    \label{fig:iteration}
\end{figure*}

\paratitle{Impact of Iterative Deliberation Turns.}
We evaluate the accuracy of DELI at different maximum numbers of iteration turns on the CMA, Algebra, and Prealgebra.
To validate the effectiveness of our designed two-stage deliberation, we also evaluate two iterative variants, \emph{i.e.,} Iterative CoT and Iterative ReAct.
As shown in Figure~\ref{fig:iteration}, the accuracy of DELI overall increases and eventually converges as the maximum number of iteration turns increases.
As a comparison, iterative variants overall do not lead to accuracy gains, even leading to a drop, which indicates that is difficult for LLM to directly discover and correct errors based on existing solutions by eliciting their comments.
In DELI, the two-stage deliberation effectively reduces model calculation errors, and considers two different perspectives of solutions to a refined solution, thus gradually improving accuracy as iterations proceed.

\begin{wraptable}{r}{0.5\textwidth}
\setlength{\tabcolsep}{3pt}
    \centering
    \caption{ExpAcc and Fail@where on a challenging subset of CARP.}
    \label{tab:hard}
    \begin{tabular}{lcccc}
    \toprule
    \multirow{2}{*}{\textbf{Methods}} &  \multirow{2}{*}{\textbf{ExpAcc}} & \multicolumn{3}{c}{\textbf{Fail@where}} \\
    \cline{3-5}
    \addlinespace[0.2em]
    & & \textbf{first} & \textbf{middle}  & \textbf{last}   \\
    \midrule
    CoT & 13.91 & 67.97 & 22.65 & 9.38 \\
    ReAct & 12.58 & 66.41 & 29.69 & 3.91 \\
    DELI & \textbf{18.90} & 60.16 & 25.78 & 14.06 \\
    \bottomrule
\end{tabular}
\end{wraptable}
\paratitle{Evaluating Intermediate Reasoning Steps.}
We evaluate intermediate reasoning steps from different prompting methods with our proposed metrics ExpAcc and Fail@where on 128 challenging problems from CARP that are incorrectly answered by all evaluated methods.
As shown in Table~\ref{tab:hard}, DELI achieves better ExpAcc than CoT and ReAct, which indicates that the method derives more correct intermediate results on the challenging subset.
Besides, Fail@where shows that DELI is less declined to generate completely wrong solutions, and has a larger percentage of near-correct solutions.
Instead of completing solutions in one go, DELI can fix errors as much as possible in the reasoning process by iterative deliberation, leading to better intermediate reasoning progress even in incorrect solutions.

\section{Conclusion}
In this paper, we proposed CARP, a computation-intensive algebra problem dataset with formulated annotation of intermediate reasoning steps for systematically evaluating LLMs in tools manipulation and math reasoning.
Based on the experiments in CARP, we found that popular LLMs with chain-of-though prompting can not solve over half of the problems in CARP, and they are more likely to make mistakes in early steps, leading to wrong answers.
To alleviate it, we proposed DELI, a new approach that can deliberate the intermediate reasoning steps with interfaces of tools.
DELI incorporated two iterative deliberation procedures to check and refine the intermediate reasoning steps of the generated step-by-step solution, from the perspectives of tool manipulation and natural language reasoning.
To verify the effectiveness of our approach, we conducted extensive experiments on CARP and 6 other computation-intensive math reasoning datasets.
Experimental results have shown that DELI outperforms baselines and can boost the performance of various CoT prompting methods.


\bibliographystyle{unsrtnat}
\bibliography{ref}

\newpage
\appendix

\section{Interface Definition}

We provide interface definitions of our tools in Table~\ref{tab:tool}.

\begin{table}[h]
\caption{Interface definitions of tools. Num. Comp., Eq. Solving, and Expr. Trans. refer to numerical computation, equation solving, and expression transformation, respectively.}
\begin{tabular}{p{1.6cm}p{4.2cm}p{7cm}}
\toprule
\textbf{Category}                         & \textbf{Interface} & \textbf{Description} \\
\midrule
\multirow{2}{*}{Num. Comp.} &    \emph{calculate}($e$) $\rightarrow v$       &         Calculate the value $v$ of $e$.    \\
 & \emph{substitute}($e$, $\{c\}$) $\rightarrow v$    & Substitute the contextual conditions $\{c\}$ into $e$. \\
\hline
\addlinespace[0.2em]
 \multirow{5}{*}{Eq. Solving} & \emph{solve\_eq}($e$) $\rightarrow , \{e'\} $   & Solve the equation $e$ to get the solution set $\{e'\}$. \\
 & \emph{solve\_ineq}($e$) $\rightarrow \{e'\}$    & Solve the inequation $e$ to get the solution set $\{e'\}$. \\
  & \emph{solve\_multi\_eq}($\{e\}$) $\rightarrow \{e'\}$    & Solve the system of equations to get the solution set $\{e'\}$. \\
 & \emph{solve\_multi\_ineq}($\{e\}$) $\rightarrow \{e'\}$    & Solve the system of inequations to get the solution set $\{e'\}$. \\
 &    \emph{partial\_solve}($e$, $u$) $\rightarrow \{e'\}$       &   Solve the equation $e$ assuming that $u$ is an unknown to get the solution set $\{e'\}$.  \\
\hline
\addlinespace[0.2em]
\multirow{4}{*}{Expr. Trans.} &    \emph{expand}($e$) $\rightarrow e'$       &         Expand $e$ to get $e'$.    \\
&    \emph{factor}($e$) $\rightarrow e'$       &         Factorize $e$ to get $e'$.    \\
&   \emph{collect}($e$, $x$) $\rightarrow e'$       &         Collect $e$ based on the symbol $x$.    \\
&    \emph{complete\_the\_square}($e$) $\rightarrow e'$       &    Complete the square of $e$ to get $e'$\\
\hline
\addlinespace[0.2em]
Thinking    &     \emph{think}($l$) $\rightarrow l'$     &     Draw a conclusion $l'$ based on the thought $l$.        \\
\bottomrule             
\end{tabular}
\label{tab:tool}
\end{table}

\section{DELI Algorithem}
The process of DELI is illustrated in Algorithm~\ref{alg:DELI}.

\begin{algorithm}[H]
    \caption{DELI algorithm}\label{alg:DELI}
    \begin{algorithmic}[1]
        \Require input problem $p$, retrieval module $R$ (optional), interfaces of tools $\{T\}$
        \State $C_R\leftarrow R(p)$     \Comment{Retrieval-augmented solution initialization}
        \State $s_n^{(0)} \leftarrow \textrm{init}(C_R,p)$ 
        \For{iteration $i \in 1 ... M$}	
            \State $s_t^{(i)} \leftarrow  \textrm{invoke}(p,s_n^{(i-1)}, \{T\})$ \Comment{Deliberation with tool interfaces}
            \State $s_n^{(i)} \leftarrow \textrm{integrate}(p, s_t^{(i)}, s_n^{(i-1)})$  \Comment{Deliberation with chain of thought}
            \If{$\textrm{equal\_answer}(s_t^{(i)},s_n^{(i)})$}    \Comment{Early stopping}
                \State break
            \ElsIf{$i\neq 0~\textrm{\textbf{and}}~\textrm{equal}(s_t^{(i-1)},s_t^{(i)})~\textrm{\textbf{and}}~\textrm{equal}(s_n^{(i-1)},s_n^{(i)})$ } 
                \State break
            \EndIf
        \EndFor
        \State \Return Last $s_t^{(i)}$
    \end{algorithmic}
\end{algorithm}


\section{Baselines}
\label{sec:baseline}
We compare our proposed DELI with several competitive prompting methods for LLMs. For all the methods, we implement them on ChatGPT.

$\bullet$ \emph{CoT prompting methods.}
We test three variants of the CoT prompting with different in-context exemplars.
\textbf{Random CoT}~\cite{Wei2022ChainOT} randomly selects exemplars from the training set.
\textbf{Complex CoT}~\cite{Fu2022ComplexityBasedPF} samples the most complex problems and their solutions as exemplars.
\textbf{Retrieval CoT} retrieves the most relevant problems and solutions from the training set as exemplars.

$\bullet$ \emph{Tool-augmented prompting methods.}
We select two tool-augmented prompting methods and implement them with our proposed interfaces on tools.
\textbf{PAL}~\cite{gao2022pal} converts the reasoning process into a Python program and executes it to get the answer.
\textbf{ReAct}~\cite{Yao2022ReActSR} invokes interfaces immediately in the reasoning process when necessary.

$\bullet$ \emph{Iterative prompting methods.}
We also compare our approach with existing iterative prompting methods and our variants.
\textbf{Learning to Program (LP)}~\cite{Guo2023LearningTP} aims to iteratively learn solutions from training sets to guide LLMs in solving similar problems based on in-context learning.
\textbf{Progressive-Hint Prompting (PHP)}~\cite{Zheng2023ProgressiveHintPI} iteratively utilizes previous answers as hints to progressive guide LLMs generating CoT solutions.
As a variant of our framework, \textbf{Iterative CoT} integrates the existing CoT solution and self-generated judgment into a refined CoT solution. Similarly, \textbf{Iterative ReAct} aims to generate better interface invocations by reviewing existing interface invocations and self-generated judgment.

\section{Implementation Details}
\label{sec:implement}
We employ OpenAI \texttt{gpt-3.5-turbo} as the solver and reasoning tool and implement the computation tool based on SymPy~\cite{Meurer2017SymPySC}.
We set the temperature to 0 and top\_p to 1 for determined outputs.
To retrieve similar problems, we train a sentence embedding model following SimCSE~\cite{Gao2021SimCSESC} to index MATH datasets and employ BM25 algorithm for the CARP dataset.
The maximum number of iteration turns is set to 3 for all datasets.
For each dataset, we specify the descriptions of interfaces that may be useful to solve the problems in prompts.

We initialize the solution with Retrieval CoT in most datasets.
For GK-Cloze and SAT-Math, we initialize the solution with Random CoT, since these datasets only provide few-shot exemplars but not training sets.
Following the settings in~\citet{Zheng2023ProgressiveHintPI}, the initial solution of PHP is from Complex CoT in subsets of MATH (Algebra, Prealgebra, CP, NT), while using the same initial solution as DELI in other datasets.

\section{Case Study}

\begin{figure*}[h]
    \centering
    \includegraphics[width=\textwidth]{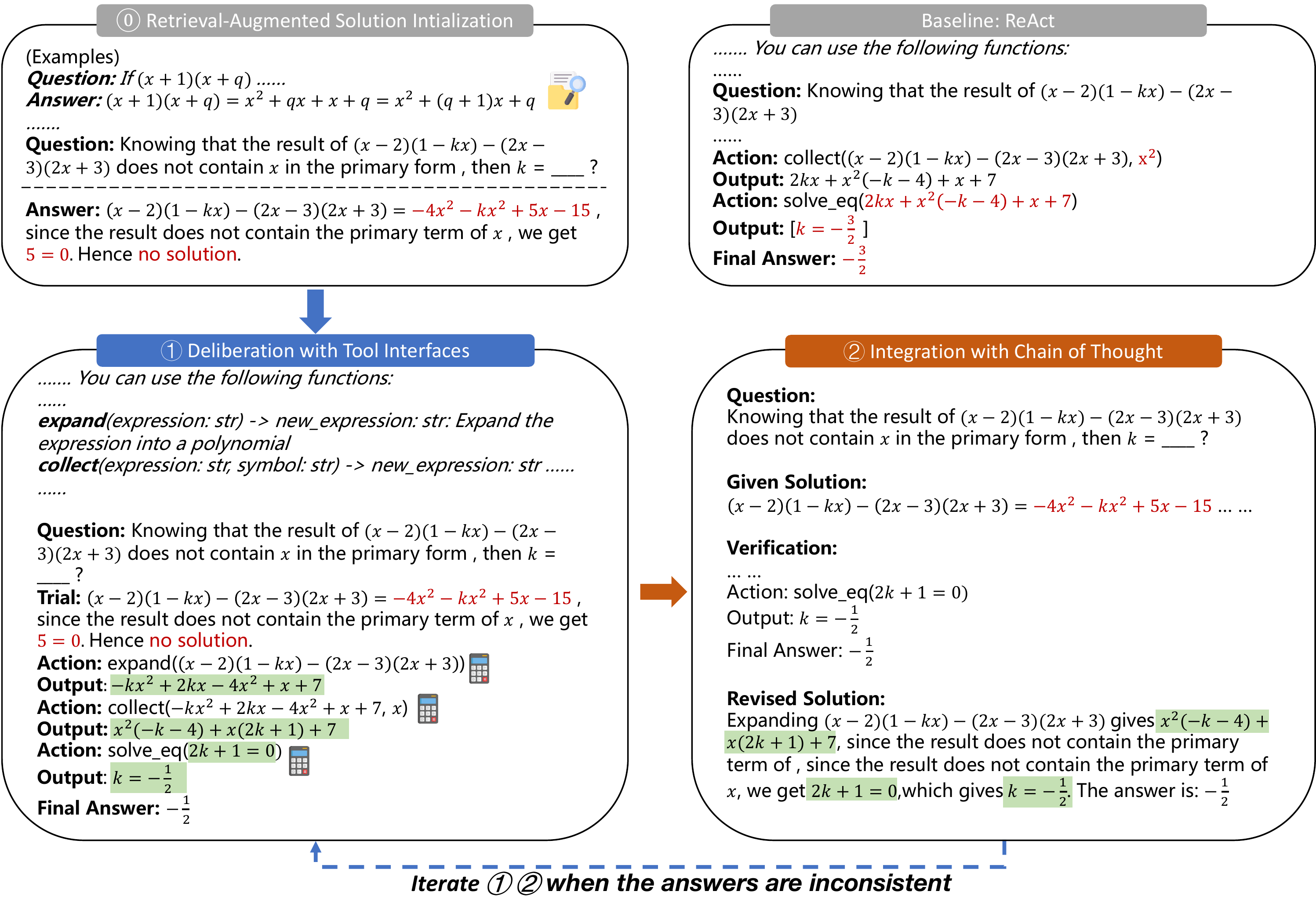}
    \caption{Case study of our method with baselines on the CARP dataset. The case is translated into English.}
    \label{fig:case}
\end{figure*}

To better present the process of DELI, we provide a case study that shows the solving process of DELI on CARP, which is shown in Figure~\ref{fig:case}.
We also report the solution of Retrieval CoT and ReAct in the figure.
It is noted that the solution of Retrieval CoT is also the initial solution in our framework.

First, both Retrieval CoT and ReAct make minor mistakes in the solving process.
Although following the correct solving idea from relevant solutions, Retrieval CoT struggles with expanding the expression, leading to an incorrect intermediate result.
ReAct fails at understanding the condition \emph{the expression does not contain the primary form of $x$}, thus collecting the expression according to a wrong term $x^2$.
Therefore, both CoT and ReAct can not solve the case individually due to the challenges of computations and reasoning.

Our method iterates over the existing solutions.
In deliberation with tool interfaces, the model reviews the existing solution, and invokes interfaces based on the ideas therein.
In this case, the model invokes interfaces \emph{expand} and \emph{collect} in a row to get the correct expanded expression with the help of tools.
Then, the model solves the equation derived from the expanded expression and gets the correct answer.

In deliberation with chain of thought, the model reviews both natural language and tool manipulation solutions from the previous iteration and generates a revised CoT solution, which fixes the computation error in the original CoT solution according to interface invocations.
In this case, due to the consistent answers between the revised CoT solution and the previous ReAct solution, the iteration terminates.
In general cases, the iteration continues until solutions converge or the answer is consistent, or the maximum number of iterations is reached.

\section{Prompts for Two-Stage Deliberation}
We list the prompts for two-stage deliberation on CARP. The prompts are translated into English.
\begin{lstlisting}[caption={Prompt for Deliberation with Tool Interfaces.}]
You are ChatGPT, a math problem solver equipped with multiple functions to tackle various math problems. While you may have access to existing problem-solving processes, there is a possibility of errors. Therefore, you need to learn the correct approaches to solve problems more efficiently. You can use the following functions: 

calculate(expression: str) -> new_expression: str: Calculate the value of the expression and return it as a string. For example, calculate("34 * 2") -> "68".
solve_eq(expression: str) -> new_expressions: list: Solve the equation expression and return the result as a list. For example, solve_eq("3 x + 4 = 1") -> ["x = -1"].
solve_ineq(expression: str) -> new_expression: str: Solve the inequality expression and return the result as a string. For example, solve_ineq("3 x + 4 < 1") -> "x < -1".
solve_multi_eq(expressions: list) -> new_expressions: dict: Solve the system of equations given by the list of expressions and return the result as a dictionary. For example, solve_multi_eq(["x + y = 2", "x - 2 y = -7"]) -> {"x": ["x = -1"], "y": ["y = 3"]}.
solve_multi_ineq(expressions: list) -> new_expression: str: Solve the system of inequalities given by the list of expressions and return the result as a string. For example, solve_multi_ineq(["x \le 2", "x \le -7"]) -> "x \le -7".
substitute(expression: str, conditions: list[str]) -> new_expression: str: Substitute the contextual conditions in the list into the expression and return the result. For example, substitute("3 x + 4", ["x = 1"]) -> "7".
expand(expression: str) -> new_expression: str: Expand the expression into a polynomial. For example, expand("(x + 1) ^ 2") -> "x ^ 2 + 2x + 1"
factor(expression: str) -> new_expression: str: Factorize the polynomial. For example, factor("x ^ 2 + 2x + 1") -> "(x + 1) ^ 2"
collect(expression: str, symbol: str) -> new_expression: str: Collect the coefficients of the corresponding powers according to the given symbol. For example, collect("a x - 5 a + x ^ { 2 } - 5 x", "x") -> "- 5 a + x ^ { 2 } + x ( a - 5 )"
partial_solve(expression: str, symbol: str) -> new_expression: str: Let the given symbol be the unknown variable and solve the linear equation expression with one variable. For example, partial_solve("x + 3 y - 3 = 0", "x) -> "x = - 3 y + 3"
think(thought: str) -> conclusion: str: Generate new conclusions based on natural language description thought. Think should only be called when the above functions are not applicable. For example, think("\sqrt{x-8} the expression inside the root is always greater than or equal to 0") -> "x-8\\ge0"

To use ChatGPT, simply provide a mathematical problem or question in LaTeX format. You can use any of the above functions to help solve the problem. Please follow the following format:

Question: The input question you must answer. This appears only once.
Trial: The problem-solving approach that can be referred to. It may contain errors, you can refer to the correct part in it.
Action: A function call, which should be one of the mentioned functions with arguments. You must only call one function in one Action.
Output: The result of the action. Every Action must be immediately followed by one and only one Output.
... (This Action/Output cycle can repeat N times.)
Final Answer: The final answer to the original input question. The answer should be numerical or LaTeX math expression. Do not use natural language in the answer
\end{lstlisting}
\begin{lstlisting}[caption={Prompt for Deliberation with Chain of Thought.}]
You are ChatGPT, a mathematical problem solver equipped with multiple functions for solving mathematical problems. You have access to both natural language problem solving processes and formal problem solving processes, but there may be errors within them. You need to learn the correct methods in order to better solve problems. You can use the following functions:

calculate(expression: str) -> new_expression: str: Calculate the value of the expression and return it as a string. For example, calculate("34 * 2") -> "68".
solve_eq(expression: str) -> new_expressions: list: Solve the equation expression and return the result as a list. For example, solve_eq("3 x + 4 = 1") -> ["x = -1"].
solve_ineq(expression: str) -> new_expression: str: Solve the inequality expression and return the result as a string. For example, solve_ineq("3 x + 4 < 1") -> "x < -1".
solve_multi_eq(expressions: list) -> new_expressions: dict: Solve the system of equations given by the list of expressions and return the result as a dictionary. For example, solve_multi_eq(["x + y = 2", "x - 2 y = -7"]) -> {"x": ["x = -1"], "y": ["y = 3"]}.
solve_multi_ineq(expressions: list) -> new_expression: str: Solve the system of inequalities given by the list of expressions and return the result as a string. For example, solve_multi_ineq(["x \le 2", "x \le -7"]) -> "x \le -7".
substitute(expression: str, conditions: list[str]) -> new_expression: str: Substitute the contextual conditions in the list into the expression and return the result. For example, substitute("3 x + 4", ["x = 1"]) -> "7".
expand(expression: str) -> new_expression: str: Expand the expression into a polynomial. For example, expand("(x + 1) ^ 2") -> "x ^ 2 + 2x + 1"
factor(expression: str) -> new_expression: str: Factorize the polynomial. For example, factor("x ^ 2 + 2x + 1") -> "(x + 1) ^ 2"
collect(expression: str, symbol: str) -> new_expression: str: Collect the coefficients of the corresponding powers according to the given symbol. For example, collect("a x - 5 a + x ^ { 2 } - 5 x", "x") -> "- 5 a + x ^ { 2 } + x ( a - 5 )"
partial_solve(expression: str, symbol: str) -> new_expression: str: Let the given symbol be the unknown variable and solve the linear equation expression with one variable. For example, partial_solve("x + 3 y - 3 = 0", "x) -> "x = - 3 y + 3"
think(thought: str) -> conclusion: str: Generate new conclusions based on natural language description thought. Think should only be called when the above functions are not applicable. For example, think("\sqrt{x-8} the expression inside the root is always greater than or equal to 0") -> "x-8\\ge0"

Follow this format:

```
Question:
The input question that you must answer. It appears only once.

Given Solution:
A natural language solution that can be used as a reference, which may contain errors. You can refer to the correct ideas in it.

Verification: Transform the original solution into a verification process that uses functions, corrects any computational errors, and simplifies the process.
Action: A function call, which must be one of the functions mentioned above and include parameters. You can only call one function in an Action.
Output: The result of an Action. Each Action must have one and only one Output following it.
(Action/Output can be repeated any number of times...)
Final Answer: The ultimate solution to the original input problem.

Revise the given solution based on the verification process:
Revise the original solution based on the computed result in the verification process. If the computed result in the verification process differs from the computed result in the original solution, the computed result in the verification process must be used as the standard.
```
\end{lstlisting}

\end{document}